\documentclass[accepted]{uai2021}

\usepackage{amsthm,amsmath,amssymb,bm}
\usepackage{appendix}
\usepackage{array}
\usepackage[american]{babel}
\usepackage{booktabs}
\usepackage{cancel}
\usepackage{enumitem}
\usepackage{fancyvrb}
\usepackage[OT1]{fontenc}
\usepackage{float}
\usepackage{graphicx}
\usepackage{macros}
\usepackage{mathtools}
\usepackage{microtype}
\usepackage{multirow}
\usepackage{nccmath}
\usepackage{tikz}
\usepackage{subcaption}
\usepackage{tcolorbox}
\usepackage{verbatim}

\usepackage[draft]{minted}
\input{mintedstyles}

\usepackage{natbib} % has a nice set of citation styles and commands
\hypersetup{
    colorlinks,
    linkcolor={blue},
    citecolor={blue},
    urlcolor={blue}
}

\usepackage{xr-hyper}
\makeatletter
\newcommand*{\addFileDependency}[1]{% argument=file name and extension
  \typeout{(#1)}
  \@addtofilelist{#1}
  \IfFileExists{#1}{}{\typeout{No file #1.}}
}
\makeatother
\newcommand*{\myexternaldocument}[1]{%
    \externaldocument{#1}%
    \addFileDependency{#1.tex}%
    \addFileDependency{#1.aux}%
}
\myexternaldocument{appendix}

\usepackage{hyperref}
\usepackage{cleveref}

\title{Active multi-fidelity Bayesian online changepoint detection}

\author[1]{Gregory W. Gundersen}
\author[1]{Diana Cai}
\author[2]{Chuteng Zhou}
\author[1]{Barbara E. Engelhardt}
\author[1]{Ryan P. Adams}
\affil[1]{%
    Department of Computer Science\\
    Princeton University
}
\affil[2]{%
    Arm ML Research Lab
}

\begin{document}

% TITLE ========================================================================
\maketitle

% ABSTRACT =====================================================================
\begin{abstract}
Online algorithms for detecting changepoints, or abrupt shifts in the behavior of a time series, are often deployed with limited resources, e.g., to edge computing settings such as mobile phones or industrial sensors. In these scenarios it may be beneficial to trade the cost of collecting an environmental measurement against the quality or ``fidelity'' of this measurement and how the measurement affects changepoint estimation. For instance, one might decide between inertial measurements or GPS to determine changepoints for motion. A Bayesian approach to changepoint detection is particularly appealing because we can represent our posterior uncertainty about changepoints and make active, cost-sensitive decisions about data fidelity to reduce this posterior uncertainty. Moreover, the total cost could be dramatically lowered through active fidelity switching, while remaining robust to changes in data distribution. We propose a multi-fidelity approach that makes cost-sensitive decisions about which data fidelity to collect based on maximizing information gain with respect to changepoints. We evaluate this framework on synthetic, video, and audio data and show that this information-based approach results in accurate predictions while reducing total cost.
\end{abstract}
% ==============================================================================

% ==============================================================================
\section{Introduction}
\label{sec:introduction}
% ==============================================================================

Sequential data are rarely stationary.
For example, a stock's volatility might increase or a text stream's topics might shift due to world events.
A changepoint is an abrupt change in the generative parameters of sequential data. The goal of changepoint detection is to discover these structural changes, and thereby partition the data into regimes
%\footnote{At a high level, changepoint detection is similar to the problem of concept drift detection~\citep{gama2004learning}.}.
Changepoint detection is a broad class of algorithms, including the classic CUSUM algorithm~\citep{page1954continuous}, hidden Markov models with a changing transition matrix~\citep{braun1998statistical}, Poisson processes with varying rates~\citep{ritov2002detection}, two-phase linear regression~\citep{lund2002detection}, and Gaussian process changepoint models~\citep{saatcci2010gaussian}. The Bayesian approach is appealing due to the ability to specify priors and represent posterior uncertainty~\citep{chib1998estimation,fearnhead2006exact,chopin2007dynamic}.
For streaming applications, exact filtering algorithms allow for online Bayesian detection of changepoints without retrospective smoothing~\citep{fearnhead2007line,adams2007bayesian}.

Many applications of online changepoint detection are in real-time settings with limited resources for sensing and computation, such as content delivery networks~\citep{akhtar2018oboe}, autonomous vehicles~\citep{ferguson2015real}, and smart home and internet-of-things devices~\citep{aminikhanghahi2018real,lee2018time, munir2019fusead}.
In such resource-constrained settings, the observations for a changepoint detector are typically environmental measurements, for example heart-rate data~\citep{villarroel2017non}.
Trading the cost of collecting these data against their quality or ``fidelity'' may be useful, depending on how these fidelities affect changepoint estimation.

For example, since scaling up neural network capacity is an effective approach to improving model performance~\citep{arora2018optimization,kaplan2020scaling,mahajan2018exploring},
a high-fidelity observation model might be a large but expensive-to-evaluate neural network. Retraining a smaller architecture or using compression algorithms such as distillation~\citep{hinton2015distilling}, quantization~\citep{gong2014compressing,hubara2017quantized}, or pruning~\citep{frankle2018lottery} could produce a low-fidelity observation model. If the output of these neural networks is the input to a changepoint detector, then the fidelity of the networks will impact the quality of changepoint detection.

In such situations, the cost of Bayesian online changepoint detection (BOCD) could be
reduced by making decisions about the fidelity of the observations.
One view of BOCD is as a model-based version of an exponentially-weighted moving average, estimating the weights from data rather than selecting them \emph{a priori}.
It determines which of the recent data matter for the current state.
This view motivates our multi-fidelity approach: if changepoints are easily identified and the data can be partitioned into stationary regimes, there is no need for expensive high-fidelity observations when BOCD's posterior confidence about changepoints is high.

In our framing of the problem, we must choose which data fidelity to use and pay a fixed cost to make this choice. In the neural network example, we can evaluate either an expensive or cheap neural network to obtain a high- or low-fidelity representation of a raw measurement. To make this choice, we propose an information-theoretic approach, similar to the active data collection strategy proposed by~\citet{mackay1992information} and to approaches used in Bayesian optimization~\citep{hernandez2014predictive}, preference learning~\citep{houlsby2012collaborative}, and Bayesian quadrature~\citep{gessner2020active}. We choose the data fidelity with maximal weighted \emph{information rate} (gain over cost) for the posterior distribution over changepoints. The weights allow modelers to specify a desired computational budget. This results in policies that use lower-fidelity data in regimes with higher posterior certainty.

\paragraph{Contributions.} First, we formulate a new version of an important problem: online changepoint detection with multiple data sources of varying cost and quality.
The task is to choose which fidelity to use at each time point to make accurate predictions while minimizing costs.
Second, we propose active selection of each datum's fidelity based on the expected informativeness of observations from each fidelity, and choose the one that maximizes the information rate for the posterior distribution over changepoints.
Finally, we demonstrate the empirical performance of our algorithm on both synthetic and real-world data.
We show that in many real-world scenarios, despite the extra step of computing information gain, our model reduces the total computational budget while maintaining good predictive accuracy.

% ==============================================================================
\section{Bayesian online changepoint detection}
\label{sec:bocd}
% ==============================================================================

We begin by reviewing the BOCD algorithm 
\citep{adams2007bayesian,fearnhead2007line}.
Our data are a contiguous sequence of observations in time,~${\boldX_{1:T} \define \{\boldx_1, \dots, \boldx_{T}\}}$ where~${\boldx_t \in \reals^D}$. 
Assume that the data can be partitioned such that, within each partition, the data are i.i.d.~\citep{barry1992product}, governed by partition-specific parameters~$\btheta$.
The transition from one partition into another results in an abrupt change from one set of parameters to another.
This transition is referred to as a changepoint.

Denote the parameters at time~$t$ as~$\btheta_t$.
In the changepoint process, these parameters are determined in one of two ways: either a changepoint has occurred at time~$t$, in which case the parameters are drawn afresh from a prior distribution~$\Pi$, or a changepoint has not occurred and the parameters are~${\btheta_t=\btheta_{t-1}}$, i.e., they stay the same.
We model the arrival of changepoints as a discrete time Bernoulli process with hazard rate~$1/\beta$, resulting in a geometric distribution over partition lengths with mean~$\beta \in \reals_{> 0}$.

In the online setting, the primary quantity of interest is the time since the last changepoint, which we refer to as the \emph{run length}.
We denote the run length at time~$t$ as~$r_t$, which takes values in the non-negative integers. Thus, a changepoint at~$t$ means ${r_t=0}$.
At time~$t$, the BOCD algorithm estimates the posterior marginal distribution over the run length~$p(r_t \given \boldX_{1:t})$. We refer to this distribution as the \emph{run-length posterior}.
Online updating of the run-length posterior is made easy via a recursion that is essentially the same as the message-passing (dynamic programming) approach to hidden Markov models~\citep{baum1966statistical,rabiner1989tutorial}:
\begin{align}
p(&r_t \given \boldX_{1:t}) \propto p(r_t, \boldX_{1:t})
\nonumber \\
&= \sum_{r_{t\!-\!1}} p(r_t, \boldx_t \given r_{t\!-\!1}, \boldX_{1:t\!-\!1}) p(r_{t\!-\!1}, \boldX_{1:t\!-\!1})
\nonumber \\
&= \sum_{r_{t\!-\!1}}
  p(r_t \given r_{t\!-\!1}, \cancel{\boldX_{1:t\!-\!1}} ) 
  p(\boldx_t \given r_t, \cancel{r_{t\!-\!1}},  \boldX_{1:t\!-\!1} )
  \nonumber \\[-10pt]
&\qquad\qquad\times  p(r_{t\!-\!1}, \boldX_{1:t\!-\!1})\nonumber\\
&= \sum_{r_{t\!-\!1}}
\underbrace{p(r_t\given r_{t\!-\!1})}_{\substack{\text{Bernoulli}\\\text{process prior}}}
\underbrace{p(\boldx_t\given r_t, \boldX_{1:t\!-\!1})}_{\substack{\text{posterior}\\\text{predictive}}}
\underbrace{p(r_{t\!-\!1},\boldX_{1:t\!-\!1}),}_{\substack{\text{previous}\\\text{estimate}}}
\label{eq:recursive_rl_estimation}
\end{align}
where the cancellations arise from Markovian assumptions we have made: 1) the probability of a changepoint at time~$t$ is independent of data before~$t$, given knowledge of~$r_{t-1}$, and 2) the predictive distribution over the data~$\boldx_t$ at time~$t$ is independent of past run lengths, given knowledge of the current run length~$r_t$.
The three terms within the sum have a convenient interpretation as the prior, the predictive distribution, and the estimated joint distribution from the previous time step. These are the only ingredients necessary for a straightforward online filtering algorithm.

The Bernoulli process prior above is in an unconventional form that represents the time since the last changepoint:
\begin{align}
    p(r_t\given r_{t-1}) &=
    \begin{cases}
    1/\beta & \text{if $r_t = 0$,}
    \\
    1- 1/\beta & \text{if $r_t = r_{t-1}+1$,}
    \\
    0 & \text{otherwise.}
    \end{cases}
\end{align}
In other words, the run length $r_t$ must either increase by one from the previous time point or drop to zero.

The construction so far has not depended on the specifics of the data-generating distribution~$P_{\btheta_t}$, which appears as a part of the posterior predictive distribution in \Cref{eq:recursive_rl_estimation}:
\begin{align}
    p(\boldx_t\given r_t \!=\! \ell, \boldX_{1:t-1}) &= \!
    \int_{\bTheta} p_{\btheta_t}(\boldx_t)\,\pi(\btheta_t\given\boldX^{(\ell)})\,
    \differential\btheta_t\,,
    \label{eq:posterior_predictive}
\end{align}
where~$p_{\btheta_t}(\cdot)$ is the probability density function associated with the distribution~$P_{\btheta_t}$, $\pi(\btheta \given \cdot)$ is the probability density function associated with the posterior distribution w.r.t. $\btheta$, and~${\boldX^{(\ell)} \define \boldX_{t-\ell:t-1}}$ denotes the most recent~$\ell$ data. This is a key property of the BOCD algorithm: conditioning on~${r_t=\ell}$ means that only the most recent~$\ell$ data need to be accounted for in the posterior distribution. When the data distribution~$P_{\btheta_t}$ is chosen to allow for a conjugate prior for~$\Pi$, then the computations necessary for the recursion are relatively simple: it is only necessary to maintain a set of sufficient statistics for each~$r_t$ hypothesis. These statistics can be easily updated via addition, and the posterior predictive is often available in closed form.
(See \citet{adams2007bayesian} for further discussion.)
When more complicated models are used, approximate inference or numerical integration are necessary.

Given the run-length posterior, we can compute a predictive distribution to make online predictions that are robust to changepoints by marginalizing out the run length, i.e., by computing a mixture of posterior predictive distributions---which are already available from the recursion---under the run-length posterior:
\begin{equation}
    p(\boldx_{t+1} \given \boldX_{1:t})
    = \bbE_{p(r_t \given \boldX_{1:t})}[
    p(\boldx_{t+1}\given r_t=\ell, \boldX^{(\ell)})
    ]\,. \label{eq:bocd_predictive}
\end{equation}
\Cref{eq:bocd_predictive} underscores the value of modeling the run-length in this construction: it provides a model-based approach to decide which data are currently relevant for predicting the next observation.
That is, the value of~$r_t$ explicitly captures the size of the current partition, i.e., what recent data share the same parameters.

The basic framework for BOCD has been extended in a number of ways, such as learning the changepoint prior~\citep{wilson2010bayesian}, adding Thompson sampling for multi-armed bandits with changing rewards~\citep{mellor2013thompson}, estimating uncertainty bounds on the number and location of changepoints~\citep{ruggieri2016exact}, and using $\beta$-divergences for robustness against outliers~\citep{knoblauch2018doubly}. While changepoint detection has been explored in the context of active data selection~\citep{osborne2010active, hayashi2019active}, to our knowledge, the BOCD framework has not been considered in multi-fidelity settings.

% ==============================================================================
\section{Multi-fidelity changepoint detection}
\label{sec:mfbocd}
% ==============================================================================

We now extend the BOCD framework to the multi-fidelity setting, referring to our algorithm as MF-BOCD. 
Our central assumption is that, at any time point~$t$, we choose the quality of our observation, with higher fidelity (lower noise) having greater cost.
We generally take this cost to be computational, but it could also be quantified in terms of resources such as money or energy.
Given the selected data fidelities, we can again recursively compute a run-length posterior 
(\Cref{sec:mf_recursive_rl_estimation}).
Given this multi-fidelity run-length posterior, the algorithm then selects
the data fidelity that maximizes a cost-sensitive information rate objective
(\Cref{sec:decision_making}).

% ==============================================================================
\subsection{Multi-fidelity posterior predictive}
% ==============================================================================

\begin{figure*}[t]
\centering
\centerline{\includegraphics[width=0.8\textwidth]{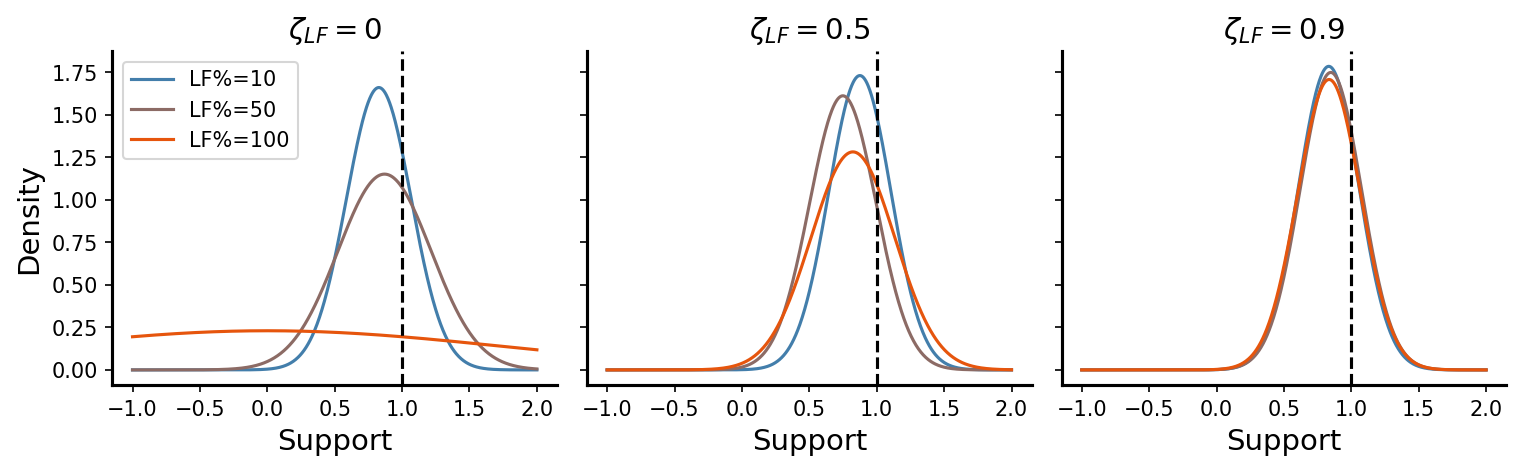}}
\caption{MF-posteriors $\pi(\theta_T \given \boldD_{1:T})$ for the Gaussian model (\Cref{sec:model_examples}) for varying low-fidelity weight $\zeta_{\textsf{LF}} \in \{0, 0.5, 0.9\}$ but fixed high-fidelity weight $\zeta_{\textsf{HF}}=1$. The data are $T=20$ i.i.d.\ samples $x_t \sim \mcN(1, 1)$. The prior is $\pi(\theta) = \mcN(0, 3)$. Within each panel, the percentage of (low-fidelity) weighted data likelihoods (LF\%) varies. When $\zeta_{\textsf{LF}}=0$ and $\text{LF\%}=100$, (left panel, orange curve) the MF-posterior reduces to the prior $\pi(\theta)$. The MF-posterior becomes more concentrated when either $\zeta_{\textsf{LF}}$ increases (right two panels) or LF\% decreases (blue curves).}
\label{fig:mf_posterior_illustration}
\end{figure*}

Again, suppose we have a distribution $P_{\btheta_t}$ and prior $\Pi$, and the task is to estimate the parameter $\btheta_t$ in the presence of changepoints. Our data are again the contiguous sequence~$\boldX_{1:T}$.

However, we now assume each observation $\boldx_t$ has an associated value $\zeta_t \in [0, 1]$, which we call the \emph{fidelity}. 
The fidelities $\boldz_{1:T} \define \{ \zeta_1, \dots, \zeta_T\}$ are non-random and take values from a set $\mcZ$. In the experiments, we only consider the case when the cardinality of $\mcZ$ is two, i.e., we only have low- and high-fidelities, but this is not a necessary restriction.
Let our sequence of observations and chosen fidelities be $\boldD_{1:T} \define \{ (\boldx_1, \zeta_1), \dots, (\boldx_T, \zeta_T)\}$. The role of the fidelity $\zeta_t$ is to re-weight the associated probability function $p_{\btheta_t}(\boldx)$ in a \emph{multi-fidelity posterior} (MF-posterior). At time $t$, the MF-posterior is:
\begin{equation}
\pi(\btheta_t \given \boldD_{1:t}) \propto \pi(\btheta_t) \prod_{i=1}^t p_{\btheta_t}(\boldx_i)^{\zeta_i}.
\end{equation}
Here, $\pi(\cdot)$ is the probability density function associated with the prior distribution $\Pi$.

Intuitively, the effect of data re-weighting on
the MF-posterior is a density that concentrates as if the contribution of $T$ samples were $\sum_{t=1}^T \zeta_t$ number of data points instead of $T$ data points.
\Cref{fig:mf_posterior_illustration} illustrates the MF-posterior of
a conjugate Gaussian model with known variance (discussed in~\Cref{sec:model_examples}). Here the data are generated from a standard normal distribution, and the MF-posterior $\pi(\theta_T \given \boldD_{1:T})$ is visualized for varying $\zeta_{\textsf{LF}}$ and fixed $\zeta_{\textsf{HF}}=1$. As $\zeta_{\textsf{LF}}$ decreases, the MF-posterior becomes less concentrated with a larger variance and increased influence from the prior.

Re-weighting terms in the likelihood has been considered under various names, such as safe Bayes~\citep{heide2020safe,grunwald2017inconsistency}, generalized posteriors~\citep{walker2001bayesian,bissiri2016general}, coarsened posteriors~\citep{miller2018robust}, and Bayesian data re-weighting~\citep{wang2017robust}. In our framing of this model, we must choose the fidelity~$\zeta_t$ of each observation~$\boldx_t$, paying a fixed cost to make this choice.

When using a member of the exponential family with a conjugate prior, one has analytical expressions of the MF-posterior and MF-posterior predictive. Let the distributions on $\boldx$ and $\btheta_t$ have the following functional forms:
\begin{align}
    p_{\btheta_t}(\boldx) \!&=\! h_1(\boldx) \exp\left\{\btheta_t^{\top} u(\boldx) - a_1(\btheta_t)\right\},
    \\
    \pi_{\bchi, \nu}(\btheta_t) \!&=\! h_2(\btheta_t) \exp\left\{\btheta_t^{\top} \bchi \!-\! \nu a_1(\btheta_t) \!-\! a_2(\bchi, \nu) \right\},
    \label{eq:exponential_family_model}
\end{align}
where, using exponential family terminology, $\btheta_t$ are now natural parameters, $u(\boldx)$ are sufficient statistics, $a_1(\cdot)$ and $a_2(\cdot, \cdot)$ are log normalizers, and $h_1(\cdot)$ and $h_2(\cdot)$ are base measures. Then the MF-posterior is
\begin{align}
\pi_{\bchi, \nu}&(\btheta_t) \prod_{i=1}^t p_{\btheta_t}(\boldx_i)^{\zeta_i} \nonumber
\\
&\propto h_2(\btheta_t) \exp\left\{ \btheta_t^{\top} \bchi_t - \nu_t a_1(\btheta_t) \right\},
\label{eq:mf_posterior}
\end{align}
where $\bchi_t = \bchi + \sum_{i=1}^t \zeta_i u(\boldx_i)$ and $\nu_t = \nu + \sum_{i=1}^t \zeta_i$. The effect of the $\zeta_i < 1$ is to down-weight the sufficient statistics of $\boldx_i$. When $\zeta_i = 1$ for all $i$, \Cref{eq:mf_posterior} reduces to the standard posterior for exponential family models.

We can now construct a multi-fidelity version of~\Cref{eq:posterior_predictive}: a posterior predictive distribution that depends on data fidelities. Let ${\boldD^{(\ell)} \define \boldD_{t-\ell:t-1}}$ denote the most recent~$\ell$ data and associated fidelities (i.e., run length~${r_t = \ell}$), and let the associated parameter estimates be:
\begin{equation}
    \bchi_{\ell} \define \bchi + \sum_{\tau=t-\ell}^{t-1} \zeta_{\tau} u(\boldx_{\tau}),
    \quad
    \nu_{\ell} \define \nu + \sum_{\tau=t-\ell}^{t-1} \zeta_{\tau}.
\end{equation}
Then the MF-posterior predictive is
\begin{align}
    p(\boldx_t &\given r_t = \ell, \zeta_t, \boldD^{(\ell)}) = \int_{\bTheta} p_{\btheta_t}(\boldx_t)^{\zeta_t} \pi(\btheta_t \given \boldD^{(\ell)}) \differential \btheta_t \nonumber
    \\
    &= h_1(\boldx_t)^{\zeta_t} \frac{\exp(a_2(\zeta_t u(\boldx_t) \!+\!\bchi_{\ell}, \zeta_t + \nu_{\ell}))}{\exp(a_2(\bchi_{\ell}, \nu_{\ell}))},
    \label{eq:mf_posterior_predictive}
\end{align}
provided~$h_1(\boldx_i)^{\zeta_i}$ induces a distribution whose normalizer we can compute. See \Cref{app:model_derivations} for a proof. This result is an extension of prior work on power posteriors for the exponential family~\citep{miller2018robust} to multiple values of powers. \Cref{eq:mf_posterior_predictive} can be interpreted as a traditional posterior predictive distribution for exponential family models but with the sufficient statistics weighted by the  fidelities. Since BOCD is amenable to fast online updates for exponential families, inference using fidelities is often no harder than using the ordinary posterior.

Note that for some multi-fidelity models, the MF-posterior~$p(\btheta_t \given r_t=\ell,\boldD^{(\ell)})$ may not have an analytic form even when~$p(\btheta_t \given \boldX^{(\ell)})$ does. In this paper, we only consider models in the exponential family, since this restriction often allows for efficient online updates. However, our approach may also extend to conditionally conjugate models. (See~\cite{miller2018robust} for a discussion.) In such settings, we could apply online variational inference to approximate predictive distributions~\citep{turner2013online}. As in standard BOCD, computing this predictive distribution without conjugate priors requires numerical approximations.

% ==============================================================================
\subsection{Multi-fidelity run-length posterior estimation}
\label{sec:mf_recursive_rl_estimation}
% ==============================================================================

To accommodate multi-fidelity observations, we must modify the online posterior estimation procedure for the run lengths.
We now condition the recursion on both the observations and data fidelities:
\begin{equation}
\begin{aligned}
p(r_t = \ell &\given \boldD_{1:t})
    \propto p(r_t ,\boldX_{1:t} \given \boldz_{1:t})
    \\
    &= \sum_{r_{t-1}}   
    p(r_t \given r_{t-1}) 
    p(\boldx_t\given r_t, \zeta_t, \boldD^{(\ell)})
    \\[-10pt]
    &\qquad\;\;\;
    \times p(r_{t-1}, \boldX_{1:t-1}\given \boldz_{1:t-1}).
    \label{eq:mbocd_recursive_rl_estimation}
\end{aligned}
\end{equation}
Similar to \Cref{eq:recursive_rl_estimation}, in the multi-fidelity case, the joint distribution of \Cref{eq:mbocd_recursive_rl_estimation} decomposes into a changepoint prior~$p(r_t\given r_{t-1})$, a predictive distribution, and the previous message. The latter two are now conditioned on fidelities.
Thus, we can efficiently update the run length posterior in a recursive manner. 

% ==============================================================================
\subsection{Examples}
\label{sec:model_examples}
% ==============================================================================

Before discussing how we choose fidelities, we demonstrate our approach with two examples of multi-fidelity models, which we use in \Cref{sec:experiments}.
To simplify notation, we ignore the run length in this section, since it only specifies which data need to be accounted for in the MF-posterior distribution. See \Cref{app:model_derivations} for more detailed derivations.

\paragraph{Multi-fidelity Gaussian.} Consider a univariate Gaussian model with known variance~$\sigma_x^2$,
\begin{equation}
    x_i \stackrel{\iid}{\sim} \mcN(\theta_t, \sigma_x^2), \quad \theta_t \sim \mcN(\mu_0, \sigma_{0}^2).
\end{equation}
The multi-fidelity likelihood is
\begin{equation}
    \prod_{i=1}^t p_{\theta_t}(x_i)^{\zeta_i} \propto \prod_{i=1}^t \exp\!\left\{-\frac{\zeta_i}{2 \sigma_x^2}(x_i - \theta_t)^2\right\},
\end{equation}
%using
and the MF-posterior is the product of~$t+1$ independent Gaussian densities, which is again a Gaussian:
\begin{align}
    \pi(\theta_t \given \boldD_{1:t})
    &\propto \mcN(\theta_t \given \mu_0, \sigma_{0}^2) \prod_{i=1}^t \mcN(x_i \given \theta_t, \sigma_x^2 / \zeta_i)
    \\
    &\propto \mcN(\theta_t \given \mu_t, \sigma_t^2),
    \label{eq:mf_gaussian}
\end{align}
where
\begin{align}
    \frac{1}{\sigma_t^2} \!=\! \frac{1}{\sigma_{0}^2} \!+\! \sum_{i=1}^t \frac{\zeta_i}{\sigma_x^2},
    \quad
    \mu_t \!=\! \sigma_t^2 \left(\frac{\mu_0}{\sigma_{0}^2} \!+\! \sum_{i=1}^t \frac{\zeta_i x_i}{\sigma_x^2} \right).
\end{align}
The MF-posterior predictive distribution can be computed by integrating out $\theta_t$. This is a convolution of two Gaussians---the posterior in~\Cref{eq:mf_gaussian} and the prior~$\pi(\theta_t)$---which is again Gaussian:
\begin{align}
    p(x_{t+1} \given \zeta_{t+1}, \boldD_{1:t}) = \mcN \!\left(x_{t+1} \,\Big|\, \mu_t, \frac{\sigma_x^2}{\zeta_{t+1}} \!+\! \sigma_t^2 \right).
\end{align}
In this example, the fidelity $\zeta_i$ has the natural interpretation of increasing the posterior variance when~${\zeta_i < 1}$. In \Cref{eq:mbocd_recursive_rl_estimation}, this has the effect that the multi-fidelity run length posterior is less concentrated. Any confidence in a changepoint is by definition lower.
 
\paragraph{Multi-fidelity Bernoulli.} Consider a Bernoulli model,
\begin{equation}
    x_i \stackrel{\iid}{\sim} \distBernoulli(\theta_t), \quad \theta_t \sim \distBeta(\alpha_0, \beta_0).
\end{equation}
The MF-posterior is proportional to a beta distribution $\pi(\theta_t \given \boldD_{1:t}) = \distBeta(\alpha_t, \beta_t)$ with parameters
\begin{equation}
    \alpha_t \define \alpha_0 + \sum_{i=1}^t \zeta_i x_i,
    \quad
    \beta_t \define \beta_0 + \sum_{i=1}^t \zeta_i (1 - x_i).
\end{equation}
The multi-fidelity posterior predictive distribution is the same as for a standard beta-Bernoulli model with $\alpha_t$ and $\beta_t$ and additional re-weighting due to $\zeta_{t+1}$:
\begin{align}
    &p(x_{t+1} \given \zeta_{t+1}, \boldD_{1:t})
    \\ \nonumber
    &= \frac{\text{B}\left( \zeta_{t+1} x_{t+1} + \alpha_t, \zeta_{t+1} (1 - x_{t+1}) + \beta_{t}
\right)}{\text{B}(\alpha_{t}, \beta_{t})},
\end{align}
where $\text{B}(\cdot,\cdot)$ is the beta function.
When $\zeta_i < 1$, the fidelity has the natural effect of discounting count observations.

% ==============================================================================
\subsection{Active fidelity selection}
\label{sec:decision_making}
% ==============================================================================

So far, we have only discussed modeling data with multiple fidelities. However, in our framing of the problem, we must actively decide the fidelity of our observation~$\boldx_t$, i.e., we must pick $\zeta_t \in \mcZ$. We propose an information-theoretic approach, similar to ideas in active data collection~\citep{mackay1992information}, Bayesian optimization~\citep{hernandez2014predictive}, preference learning~\citep{houlsby2012collaborative}, and Bayesian quadrature~\citep{gessner2020active}. We propose maximizing the weighted information rate of the multi-fidelity run length distribution. After observing $\boldD_{1:t-1}$ observations and fidelities, our current information about $r_t$ is the Shannon entropy~$\bbH[p(r_t \given \boldD_{1:t-1})]$. 
Since we must choose a fidelity without observing $\boldx_t$, we want to choose the one that minimizes the expected entropy with respect to the predictive distribution in \Cref{eq:bocd_predictive}. Thus, we choose the fidelity that maximizes the information gain of the run length posterior. The \emph{utility} of $\zeta_t$ is therefore
\begin{equation}
    \mcU(\zeta_t) = \bbH[r_t \given \boldD_{1:t-1}] - \bbE_{\boldx_t}[\bbH[r_t\given \boldD_{1:t-1}, \boldx_t, \zeta_t]].
    \label{eq:mi_rx}
\end{equation}
At time $t$, the left term in \Cref{eq:mi_rx} is easy to compute, since we have already computed the posterior distribution $p(r_{t-1} \given \boldD_{1:t-1})$. We simply roll our estimation forward in time according to the changepoint process and without conditioning on new data. Furthermore, this value is the same for all fidelities, and therefore an equivalent formulation is to minimize the expected run length entropy, the right term in \Cref{eq:mi_rx}. This entropy term is easy to compute because it is with respect to a discrete distribution that we can estimate at time $t$. The expectation is with respect to the predictive distribution (\Cref{eq:bocd_predictive}) and must be approximated in general.

However, we are not interested in the fidelity that just maximizes information gain regardless of cost. If this were the case, we would simply always use the highest fidelity. Let $\lambda(\zeta_t)$ denote the cost of fidelity $\zeta_t$. In general, $\lambda(\cdot)$ could be a function of the input domain, but here we assume it is a scalar constant that is known, e.g., wall-time, energy usage, or floating point operations. Then the \emph{information rate} of fidelity $\zeta_t$ at time $t$ is~$\alpha(\zeta_t) \define \mcU(\zeta_t) / \lambda(\zeta_t)$.
However, given the interaction of fixed costs and estimated fidelities, it is possible that the maximum information rate is always achieved using the highest (or lowest) fidelity. In this case, we may still want some amount of low-fidelity (or high-fidelity) usage depending on dataset size and computational budget. To address this, consider arbitrary weights $w(\zeta_t) \geq 0$. Our decision rule is then: use fidelity $\zeta_t^{\star}$ that maximizes the weighted information rate:
\begin{equation}
    \zeta_t^{\star} \define \argmax_{\zeta_t \in \mcZ} w(\zeta_t) \alpha(\zeta_t).
    \label{eq:decision_rule}
\end{equation}
Note that the weights can be tuned on held-out data to achieve a desired expected budget. Introducing weights is useful because we do not lose $\lambda(\zeta_t)$, which may represent an interpretable quantity such as floating point operations.

We considered alternative decision rules to~\Cref{eq:decision_rule}. For example, in scenarios with just two fidelities (low and high), we explored a decision rule that picked the low-fidelity datum when the absolute difference in information gains was less than some \emph{margin} hyperparameter. However, empirically, this resulted in frequent switching between fidelities since the two information gains were often quite close in value. We found that information rate was more stable because it requires a more significant change in information gain to induce a switch. See~\Cref{app:alternative_decision_rule} for a discussion and additional results.

% ==============================================================================
\subsection{Practical considerations}
% ==============================================================================

\paragraph{Analyzing costs.} Since we are motivated by real-time decision-making, a sensible question is whether our decision-making algorithm is cheaper than using only high-fidelity observations.
Here, we give a complete example of the cost for the beta-Bernoulli model. Since the predictive distribution is easy to work with, a useful reformulation of \Cref{eq:mi_rx} is
\begin{equation}
    \mcU(\zeta_t) = \bbH[\boldx_t \given \boldD_{1:t-1}] - \bbE_{r_t}[\bbH[\boldx_t \given \boldD_{1:t-1}, r_t, \zeta_t]],
    \label{eq:mi_xr}
\end{equation}
which uses the symmetry of information gain. 
At time $t$, the cost in floating point operations (flops) of computing \Cref{eq:mi_xr} is~$32t + 1$ flops. The cost grows linearly with time because computing information gain requires summing over the run length posterior $p(r_t \given \boldD_{1:t-1})$, and the support of this distribution grows linearly with time. However, \citet{fearnhead2007line} proposed an optimal resampling algorithm, similar to particle filtering,
that enables efficient approximate inference. This allows for a fixed cost to compute information gain. For example, with 10,000 particles, computing the information gain for the Bernoulli model requires 0.32 million flops. For comparison, consider MobileNets, which are a class of efficient neural networks designed for mobile and embedded vision applications~\citep{howard2017mobilenets}. The smallest reported MobileNet requires 41 million multi-adds (82 million flops). Thus, computing the beta-Bernoulli information gain twice (when the cardinality of $\mcZ$ is 2) is 140 times cheaper than evaluating the smallest MobileNet, while still using 10,000 particles in the run length posterior estimation.

\paragraph{Estimating fidelity $\zeta_t$.} A second practical consideration is estimating $\zeta_t$. In the Gaussian case with known variance $\sigma_x^2$, we can estimate $\zeta_t / \sigma_x^2$ using the sample variance of held-out data and then calculate the value for $\zeta_t$. In the Bernoulli case, we use model accuracy as a proxy for $\zeta_t$. For example, if a binary classifier has a true positive rate of 90\%, we treat an observation of 1 as a 0.9 using $\zeta_t=0.9$.

% ==============================================================================
\section{Experiments}
\label{sec:experiments}
% ==============================================================================

In this section, we empirically evaluate our algorithm on synthetic, video, and audio data, and compare performance of MF-BOCD against BOCD using only low- or high-fidelity data, as well as a randomized baseline. Please see \Cref{app:didactic_code} for didactic code and the repository for a complete implementation.\footnote{\url{https://github.com/princetonlips/mf-bocd}}

To evaluate our framework, we define two metrics. Let $\bar{\boldX}_{1:T} \define \{\bar{\boldx}_1, \dots, \bar{\boldx}_T\}$ denote the mean of the predictive distribution, \Cref{eq:bocd_predictive}, of BOCD or MF-BOCD for all time points. Then the reported mean squared error (MSE) is between $\bar{\boldX}_{1:T}$ from the evaluated model and $\bar{\boldX}_{1:T}$ from BOCD using only high-fidelity data. Now let $\boldR_{1:T}$ denote a lower triangular matrix denoting the run length posterior at all time points. The $L_1$ distance is between $\boldR_{1:T}$ from the evaluated model and $\boldR_{1:T}$ from BOCD using only high-fidelity data. In other words, we compare the evaluated model to the best it could have done in practice.

As a baseline, we compare MF-BOCD with a model that randomly switches between fidelities and which uses roughly the same percentage of high-fidelity data as MF-BOCD. For the random switching model, the decision to use low-fidelity data was based on the outcome of a Bernoulli random variable with bias equal to the percentage of low-fidelity data used by MF-BOCD, normalized to $[0,1]$.  This comparison isolates the question: is it \emph{when} a multi-fidelity model uses high-fidelity data that improves performance or just the presence of high-fidelity data at all?
\begin{figure*}[t]
\centering
\centerline{\includegraphics[width=\textwidth]{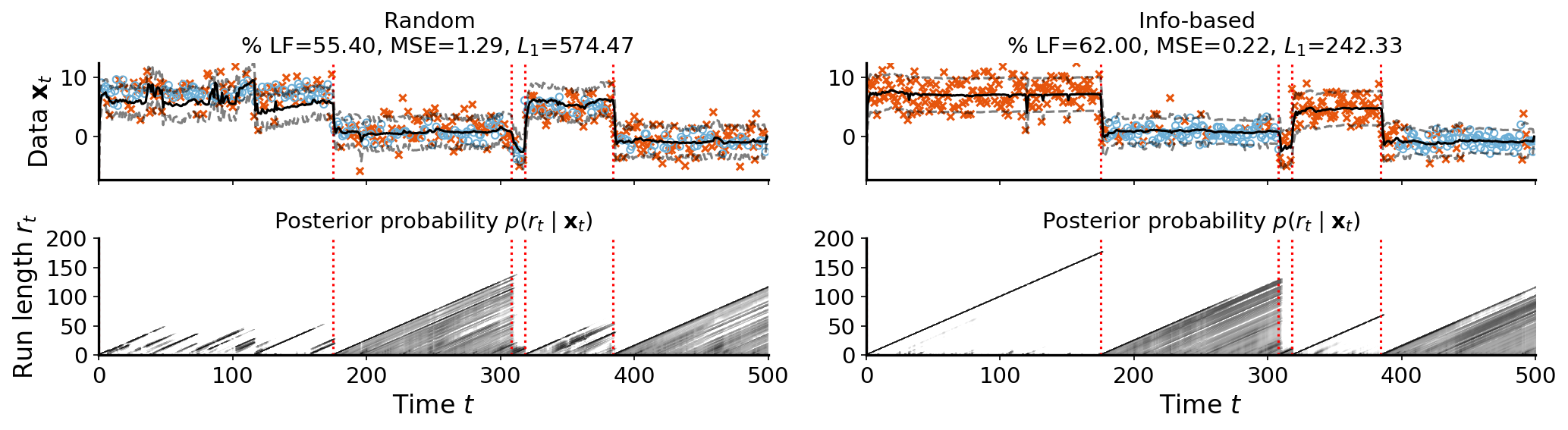}}
\caption{Comparison of two multi-fidelity models. Orange x marks and blue circles denote low- and high-fidelity data respectively. (Left two panels) A multi-fidelity model with random switching between fidelities. The probability of switching to low-fidelity data was chosen to be the fraction of low-fidelity observations used in the experiment in the right column. (Right two panels) MF-BOCD: a multi-fidelity model that actively selects the fidelity based on information rate.}
\label{fig:mi_illustration}
\end{figure*}
%
% ==============================================================================
\subsection{Numerical experiments}
\label{sec:numerical_experiments}
% ==============================================================================

The purpose of these experiments is to demonstrate that information rate is a useful decision rule and to build intuition about the model's behavior in a controlled setting. Consider a synthetic univariate signal with two fidelities. We assume data are i.i.d.\ Gaussian within each partition, and we use the Gaussian multi-fidelity model described in \Cref{sec:model_examples}.
When a changepoint occurs, the parameter $\theta_t$ is drawn from a prior $\mcN(1, 3)$. The data is then drawn from a distribution $x_t \sim \mcN(\theta_t, \zeta/\sigma_x^2)$ where~$\sigma_x^2=1$. Our fidelities are from the set $\mcZ = \{\zeta_{\textsf{HF}}, \zeta_{\textsf{LF}}\}$. We set the higher fidelity to $\zeta_{\textsf{HF}} = 1$ and the lower fidelity to $\zeta_{\textsf{LF}} = 1/2$. Thus, low-fidelity data have twice the variance. Costs are arbitrary in this setting, and we set them to $\lambda(\zeta_{\textsf{HF}}) = 2$ and $\lambda(\zeta_{\textsf{LF}}) = 1$. We simulated the data using $T = 500$ observations with a changepoint prior with $1/\beta = 1/100$.

This experiment illustrates information rate as a decision rule as described in \Cref{sec:decision_making}. In regions in which the model is confident about the run length posterior, low-fidelity data are preferred because both fidelities provide sufficient information.
However, when the model is uncertain about the run length posterior, the high-fidelity observations are preferred (\Cref{fig:mi_illustration}). In contrast to information-based switching, the multi-fidelity model with random switching has both higher MSE and $L_1$ metrics.This suggests that while just using some high-fidelity data is useful, choosing when to use that high-fidelity data can improve performance.
While this result is illustrative, we also include two randomized ablation experiments in~\Cref{app:ablations}.

% ==============================================================================
\subsection{Cambridge video data}
\label{sec:camvid}
% ==============================================================================

The numerical experiments provide a useful illustration of the role of information gain in a controlled setting. However, the fidelities and costs are contrived. In this section, we present a complete example of MF-BOCD with observation models and associated costs for the purpose of real-time detection of changepoints in streaming video data.

The Cambridge-driving Labeled Video Database (CamVid) is a collection of over ten minutes of video footage with object class semantic labels from 32 classes~\citep{brostow2009semantic}. The videos have been manually labeled at 1 frame per second, for just over 700 images. Each frame is $320 \times 480$ pixels. For observation models, we used pretrained V3 MobileNets~\citep{howard2017mobilenets,howard2019searching}. The high-fidelity model is larger and more accurate (\Cref{tab:mobile_micro_nets} in \Cref{app:experimental_details}).

\begin{figure*}[t!]
\centering
\centerline{\includegraphics[width=\textwidth]{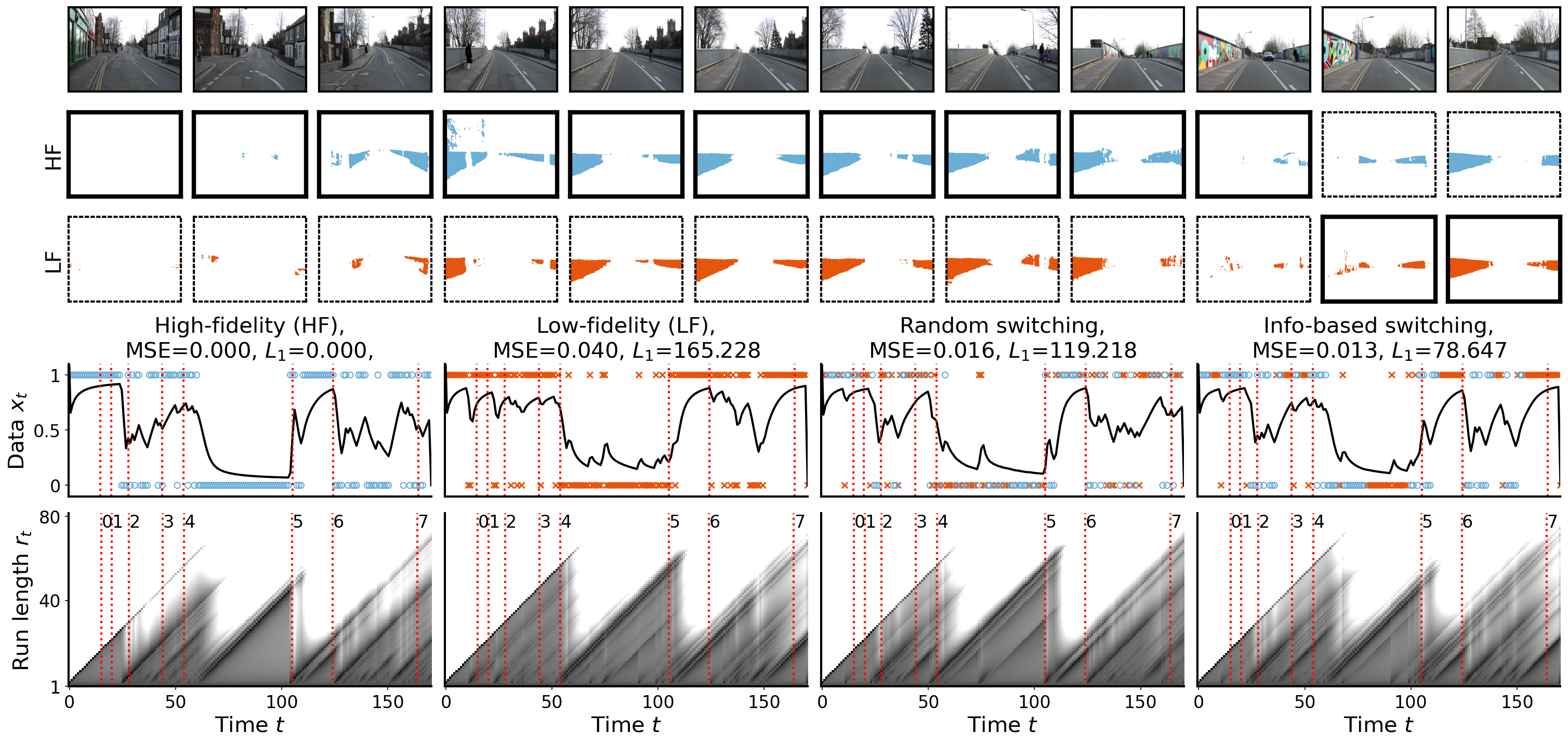}}
\caption{MF-BOCD on CamVid video stream. (Top three rows) A sequence of video frames as a camera-mounted vehicle approaches a bridge with fences on either side. The high- and low-fidelity masks are shown in middle and bottom rows respectively. A solid black frame indicates which fidelity was used by MF-BOCD. (Main center row) Binarized output from MobileNets for high-, low-, and multi-fidelity models. The solid black lines are predictive means. (Bottom row) Run length posteriors along with changepoints manually labeled from the groundtruth masks.% any trade-offs as to relative cost of HF vs MF?
}
\label{fig:camvid}
\end{figure*}

The output of each observation model is a segmentation mask, which we converted to a binary signal depending on whether or not a given class is in the image. In particular, we used the ``fence'' signal because fences go in and out of the frame but typically remain in a sequence of frames for a brief period. We then fit the multi-fidelity Bernoulli model (\Cref{sec:model_examples}) to the CamVid test set. We used the predictive version of information gain, \Cref{eq:mi_xr}. We arbitrarily set the low-fidelity model's cost to 1 and the high-fidelity model's cost as function of that, $36.7/19.5 \approx 1.9$, using the number of flops (in billions) as a proxy for cost (\Cref{tab:mobile_micro_nets}). The high-fidelity model used $\zeta_{\textsf{HF}} = 1$. The low-fidelity model's fidelity is a function of the difference in mean intersection-over-union for each model, $\zeta_{\textsf{LF}} = 1 - (0.723 - 0.674) \approx 0.95$.

We found that the output of low- and high-capacity neural networks were a reasonable proxy for low- and high-fidelity data. Standard BOCD using only high-fidelity observations estimates a run-length posterior that captures more groundtruth changepoints and has a predictive mean with smaller MSE and $L_1$ distance than BOCD using only low-fidelity data. The multi-fidelity model's decision rule weights were tuned to approximate total computational cost of $50\%$ low-fidelity data using cross-validation data, and the randomized approach flips a fair coin to choose the data fidelity. On test data, MF-BOCD estimated a run length posterior that still closely matched the high-fidelity run-length posterior (\Cref{fig:camvid}). The information-based approach results in a better predictive mean (MSE) and better run length posterior estimation ($L_1$ distance) than both the low-fidelity and randomized versions.

Finally, we estimated the computational cost of MF-BOCD relative to baselines. With roughly 50\% low-fidelity data, the costs in billions of flops for MF-BOCD was 4827, for BOCD using just low-fidelity data was 3333, and for BOCD using just high-fidelity data was 6303. The cost of decision making was marginal, requiring 0.00046 billion flops (\Cref{app:experimental_details}). As this calculation demonstrates, making a decision between high- and low-capacity neural networks can be significantly cheaper than evaluating either model. So while random usage of low-fidelity data is a reasonable approach to lowering the computational budget, decision-making can improve inference and predictions with marginal added cost.

\begin{figure}[t]
\centering
\centerline{\includegraphics[width=\columnwidth]{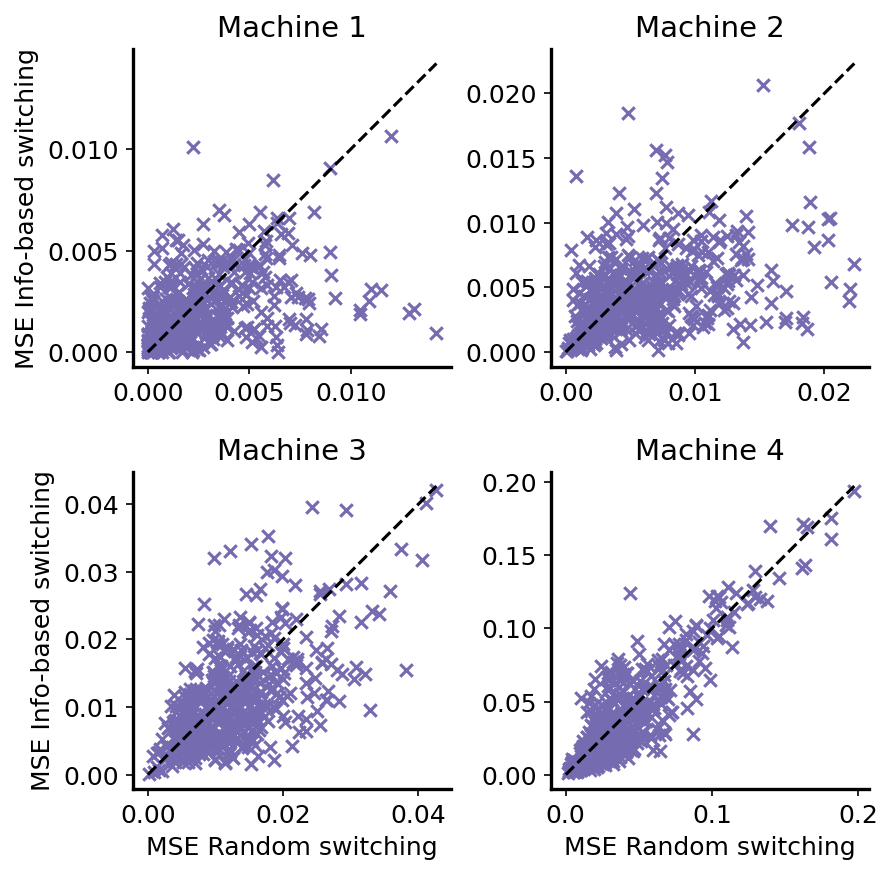}}
\caption{Comparison of information-based vs. random switching on the MIMII dataset. Under the line is better for MF-BOCD. See~\Cref{tab:mimii} for means and standard errors.}
\label{fig:mimii_visualization}
\end{figure}

% ==============================================================================
\subsection{MIMII audio data}
\label{sec:mimii}
% ==============================================================================

Next, we evaluated MF-BOCD on the sound dataset for Malfunctioning Industrial Machine Investigation and Inspection~\citep[MIMII,][]{purohit2019mimii}. The raw data are 10-second audio clips recorded from 4 different industrial machines (slide rails in this experiment) during either normal or anomalous operation. For example, anomalous conditions might involve rail damage, a loose belt, or no grease. The high-fidelity observation model is a depth-wise separable convolutional neural network~\citep[MicroNets,][]{banbury2020micronets}. The low-fidelity observation model is a two-layer fully-connected neural network. Both models take frames of log-Mel spectrograms of audio signals as inputs and return an anomaly score as output. They were pretrained on audio clips of normal behavior. Then each 10-second test set clip was converted to 14 anomaly scores using these observation models. The anomaly score is a number between 0 and 1, with 0 indicating normal. We thresholded the anomaly scores to produce binary labels. We picked machine- and model-specific thesholds using ROC curves. (See \Cref{app:experimental_details} for details.)

To randomly generate audio files with changepoints, we sampled a sequence of Bernoulli random variables $\boldy_{1:T}$. Then for each $y_t$, we chose a normal (anomalous) audio clip uniformly at random with replacement if~$y_t = 0$ ($y_t = 1$). We converted clips to low- (high-) fidelity data by evaluating the low- (high-) neural network and computing the median anomaly score for that clip. As in~\Cref{sec:camvid}, we used a Bernoulli model with $\zeta_{\textsf{HF}}=1$ and $\zeta_{\textsf{LF}}$ set to the low-fidelity model's true positive rate relative to the high-fidelity model. For each machine, we randomly generated 500 datasets with changepoints and computed the MSE and $L_1$ distances for low-fidelity BOCD and for MF-BOCD with both random and information-based switching. We found that the information-based approach to switching had lower MSE and $L_1$ distance than BOCD using just low-fidelity data and had better performance than randomized switching on the first three machines (\Cref{tab:mimii}). An interesting negative result is that MF-BOCD does not do significantly better than random on machine 4. We hypothesize that this is due to the poor quality of the low-fidelity observation model, which has an AUC $< 0.5$ (\Cref{fig:mimii_threshold_estimation}, Appendix). With these data, MF-BOCD is making hard decisions (argmax) with bad information. And in general, a randomized approach can sometimes do well (\Cref{fig:mimii_visualization}). An interesting direction for future work would be to soften the decision rule via sampling, perhaps controlled by a temperature.

As in the CamVid experiments, we found that the total cost of decision-making was marginal; the neural network costs dominated the calculations (\Cref{tab:mimii}). Thus, MF-BOCD offers a useful way to trade off detection accuracy for computational savings.

\begin{table}[t]
\caption{Comparison between low-fidelity BOCD (LF), random switching (RN), and MF-BOCD (IG). Mean and two standard errors were computed over 500 randomly generated MIMII datasets with changepoints, using the method described in the text. Cost is in millions of flops. Bold numbers indicate statistically significant using 95\% confidence intervals. \%LF is the percentage of low-fidelity data used by both multi-fidelity models, RN and IG. The reported \%LF is the average across all datasets.}
\label{tab:mimii}
\resizebox{\columnwidth}{!}{%
\begin{tabular}{l|l|c|c|c|c}
\toprule
\midrule
& & Machine 1 & Machine 2 & Machine 3 & Machine 4 \\
\midrule
%
% \parbox[t]{2mm}{\multirow{3}{*}{\rotatebox[origin=c]{90}{MSE}}}
\multirow{3}{*}{MSE}
& LF & 0.0060 (0.0004) & 0.0195 (0.0008) & 0.0347 (0.0012) & 0.1743 (0.0042) \\
& RN & 0.0026 (0.0002) & 0.0063 (0.0004) & 0.0126 (0.0006) & 0.0411 (0.0028) \\
& IG & \textbf{0.0020} (0.0002) & \textbf{0.0045} (0.0003) & \textbf{0.0112} (0.0006) & 0.0393 (0.0030) \\
\midrule
%
% \parbox[t]{2mm}{\multirow{3}{*}{\rotatebox[origin=c]{90}{$L_1$}}}
\multirow{3}{*}{$L_1$}
& LF & 101.87 (3.28) & 167.73 (3.61) & 192.49 (4.02) & 242.85 (4.63) \\
& RN & 57.61 (3.14) & 97.79 (3.66) & 132.06 (3.63) & 178.86 (4.97) \\
& IG & 61.79 (3.02) & 92.98 (3.65) & 130.17 (3.88) & 173.27 (4.95) \\
\midrule
%
% \parbox[t]{2mm}{\multirow{3}{*}{\rotatebox[origin=c]{90}{Ops}}}
\multirow{3}{*}{Ops}
& LF & 100 & '' & '' & '' \\
& RN & 14447.58 & 16109.38 & 12867.76 & 13357.11 \\
& IG & 14448.22 & 16110.02 & 12868.40 & 13357.74 \\
& HF & 24940 & '' & '' & '' \\
\midrule
%
% \parbox[t]{1mm}{\multirow{1}{*}{\rotatebox[origin=c]{90}{LF \%}}}
\%LF & & 42 & 36 & 48 & 46 \\
\bottomrule
\end{tabular}
}
\end{table}

% ==============================================================================
\section{Discussion}
% ==============================================================================

We have extended Bayesian online changepoint detection to the multi-fidelity setting in which observations have associated fidelities and costs. We found that choosing the data fidelity based on maximal information rate with respect to the run-length posterior yields interpretable policies that lower computational costs while still maintaining good performance in terms of parameter and run-length posterior estimation. In simple models, decision-making is cheap relative to the cost of evaluating even tiny neural networks designed for commodity microcontrollers. Flops savings translate to energy savings~\citep{banbury2020micronets}, which is crucial for resource-constrained applications.

While we focus on the online and resource-constrained setting, this framework could be extended to scenarios in which observations take a long time to compute, such as changepoint detection in protein-folding~\citep{fan2015identifying} or engineering design~\citep{robinson2008surrogate}. In such settings, expensive approximations of the posterior predictive distribution or information gain may be tolerable, as well as retrospective smoothing of the run-length distributions.

Alternative decision rules should also be explored, as these will induce different policies. \citet{gessner2020active} discuss how any monotonic transformation of \Cref{eq:decision_rule} gives rise to the same policy because the global maximum is the same even if the value at that maximum is not. However, this is not necessarily true after dividing the decision rules by costs. Furthermore, a probabilistic decision rule might be useful in scenarios where the difference between low- and high-fidelity observation models is marginal.

% ==============================================================================
\subsection*{Acknowledgements}
% ==============================================================================

We thank Paul Whatmough and Igor Fedorov for helpful conversations on ML for resource-constrained devices. 
B.E.\ Engelhardt and G.W.\ Gundersen received support from a grant from the Helmsley Trust, a grant from the NIH HTAN Research Program, NIH NHLBI R01 HL133218, and NSF CAREER AWD1005627.
D. Cai was supported in part by a Google Ph.D. Fellowship in Machine Learning.  R.P.\ Adams was supported in part by NSF IIS-2007278.

% \nocite{*}
\bibliography{main-uai2021}

\clearpage
\appendix
\begin{appendices}

\onecolumn
\title{Active multi-fidelity Bayesian online changepoint detection\\Supplementary material}
\maketitle
% \vspace*{-4cm}

% ==============================================================================
\section{Model derivations}\label{app:model_derivations}
% ==============================================================================

% ==============================================================================
\subsection{MF-posterior predictive for exponential family models}
% ==============================================================================

In multi-fidelity BOCD, we desire the posterior predictive distribution conditioned on the run length,
\begin{equation}
    p(\boldx_t \given r_t = \ell, \zeta_t, \mathbf{D}_{t-\ell:t-1}).
\end{equation}
Assume this is an exponential family model with the following likelihood and and prior density functions:
\begin{align}
    p_{\btheta_t}(\boldx) &= h_1(\boldx) \exp\left\{\btheta_t^{\top} u(\boldx) - a_1(\btheta_t)\right\},
    \\
    \pi_{\bchi, \nu}(\btheta_t) &= h_2(\btheta_t) \exp\left\{\btheta_t^{\top} \bchi - \nu a_1(\btheta_t) - a_2(\bchi, \nu) \right\}.
\end{align}
See \Cref{sec:mfbocd} or \Cref{eq:exponential_family_model} for a description of these terms. We introduce the following notation to denote the data and parameter estimates for the previous $\ell$ observations, associated with the run length hypothesis $r_t = \ell$:
\begin{equation}
    \mathbf{D}^{(\ell)} \define \mathbf{D}_{t-\ell:t-1},
    \quad
    \bchi_{\ell} \define \bchi + \sum_{\tau=t-\ell}^{t-1} \zeta_{\tau} u(\boldx_{\tau}),
    \quad
    \nu_{\ell} \define \nu + \sum_{\tau=t-\ell}^{t-1} \zeta_{\tau}.
\end{equation}
Then the posterior predictive is
\begin{align}
    &p(\boldx_t \given r_t = \ell, \zeta_t, \mathbf{D}^{(\ell)})
    \\
    &= \int_{\bTheta} p_{\btheta}(\boldx_t)^{\zeta_t} \pi_{\bchi_{\ell}, \nu_{\ell}}(\btheta) \differential \btheta
    \\
    &= \int_{\bTheta} [h_1(\boldx_t)]^{\zeta_t} \exp\left\{\btheta^{\top} \zeta_t u(\boldx_t) - \zeta_t  a_1(\btheta) \right\} 
    \\
    &\quad\;\; h_2(\btheta) \exp\left\{ \btheta^{\top} \bchi_{\ell} - \nu_{\ell} a_1(\btheta) - a_2(\bchi_{\ell}, \nu_{\ell}) \right\} \differential\btheta
    \\
    &= [h_1(\boldx_t)]^{\zeta_t} \frac{\int_{\bTheta} h_2(\btheta) \exp\left\{\btheta^{\top} \left[ \zeta_t u(\boldx_t) + \bchi_{\ell} \right] - a_1(\btheta) \left[ \zeta_t + \nu_{\ell} \right] \right\} \differential\btheta}{\exp\left\{a_2(\bchi_{\ell}, \nu_{\ell})\right\}}
    \\
    &\stackrel{\star}{=} [h_1(\boldx_t)]^{\zeta_t} \frac{\exp\left\{a_2(\zeta_t u(\boldx_t) + \bchi_{\ell}, \zeta_t + \nu_{\ell})\right\}}{\exp\left\{a_2(\bchi_{\ell}, \nu_{\ell})\right\}}
    \\
    &= [h_1(\boldx_t)]^{\zeta_t} \exp\left\{a_2(\zeta_t u(\boldx_t) + \bchi_{\ell}, \zeta_t + \nu_{\ell}) - a_2(\bchi_{\ell}, \nu_{\ell})\right\}
\end{align}
Step $\star$ follows from the previous line because we know the normalizer for the integral. This result is similar to the result on power posteriors for the exponential family~\citep{miller2018robust}. However, our approach requires multiple values of powers, which represent data fidelities.

% ==============================================================================
\subsection{Multi-fidelity Gaussian model}
% ==============================================================================

To simplify notation, we ignore the run length in this section, since it only specifies which data need to be accounted for in the MF-posterior distribution. Consider a univariate\footnote{This result straightforwardly extends to the multivariate Gaussian.} Gaussian model with known variance.
\begin{equation}
    x_i \stackrel{\iid}{\sim} \mcN(\theta_t, \sigma_x^2), \quad \theta_t \sim \mcN(\mu_{0}, \sigma_{0}^2).
\end{equation}
The multi-fidelity likelihood is
\begin{align}
    \prod_{i=1}^t p_{\theta_t}(x_i)^{\zeta_i}
    &= \prod_{i=1}^t \left[ \frac{1}{\sqrt{2\pi} \sigma_x^2} \exp\left\{-\frac{1}{2\sigma_x^2}(x_i - \theta_t)^2\right\} \right]^{\zeta_i}
    \\
    &\propto \prod_{i=1}^t \exp\left\{-\frac{\zeta_i}{2 \sigma_x^2}(x_i - \theta_t)^2\right\}
\end{align}
When $\zeta_i < 1$, the variance of $\mcN(x_i \given \sigma_x^2 / \zeta_i)$ increases, and the fidelity hyperparameter has the natural interpretation of increasing the variance of our model.

The multi-fidelity posterior is the product of $t+1$ independent Gaussian densities, which is itself Gaussian:
\begin{align}
    \pi(\theta_t \mid \boldD_{1:t})
    &\propto \mcN(\theta_t \given \mu_0, \sigma_{0}^2) \prod_{i=1}^t \mcN(x_i \given \theta_t, \sigma_x^2 / \zeta_i)
    \\
    &\propto \mcN(\theta_t \given \mu_t, \sigma_t^2),
\end{align}
where
\begin{align}
    \frac{1}{\sigma_t^2} = \frac{1}{\sigma_{0}^2} + \sum_{i=1}^t \frac{\zeta_i}{\sigma_x^2},
    \quad
    \mu_t = \sigma_t^2 \left(\frac{\mu_0}{\sigma_{0}^2} + \sum_{i=1}^t \frac{\zeta_i x_i}{\sigma_x^2} \right).
\end{align}
The MF-posterior predictive can be computed by integrating out $\theta_t$. This is a convolution of two Gaussians, the posterior in~\Cref{eq:mf_gaussian} and the prior $\pi(\theta) = \mcN(\theta \given \mu_0, \sigma_0^2)$, which is again Gaussian:
\begin{align}
    p(x_{t+1} \given \zeta_{t+1}, \mathbf{D}_{1:t})
    &= \int_{\Theta} [\mcN(x_{t+1} \given \theta_t, \sigma_x^2)]^{\zeta_{t+1}} \mcN(\theta_t \given \mu_t, \sigma_t^2) \differential\theta_t
    \\
    &= \mcN\left(x_{t+1} \given \mu_t, \frac{\sigma_x^2}{\zeta_{t+1}} + \sigma_t^2 \right).
\end{align}

With a single fidelity and $\zeta=1$, this results reduces to the standard result for Gaussian models with known variance~\citep{murphy2007conjugate}.

% ==============================================================================
\subsection{Multi-fidelity Bernoulli model}
% ==============================================================================

To simplify notation, we ignore the run length in this section, since it only specifies which data need to be accounted for in the MF-posterior distribution. Consider a beta-Bernoulli model
\begin{equation}
    x_i \stackrel{\iid}{\sim} \distBernoulli(\theta_t), \quad \theta_t \sim \distBeta(\alpha_0, \beta_0).
\end{equation}
The multi-fidelity likelihood is
\begin{align}
    \prod_{i=1}^t p_{\theta_t}(x_i)^{\zeta_i}
    &= \prod_{i=1}^t \left[ \theta_t^{x_i} (1 - \theta_t)^{1 - x_i}\right]^{\zeta_i}
    \\
    &= \prod_{i=1}^t \theta_t^{\zeta_i x_i} (1 - \theta_t)^{\zeta_i (1 - x_i)}.
\end{align}
Therefore the MF-posterior is
\begin{align}
    \pi(\theta_t) \prod_{i=1}^t p_{\theta_t}(x_i)^{\zeta_i}
    &\propto \frac{1}{\text{B}(\alpha_0, \beta_0)} \theta_t^{\alpha_0-1} (1 - \theta_t)^{\beta_0 - 1} \prod_{i=1}^t \theta_t^{\zeta_i x_i} (1 - \theta_t)^{\zeta_i (1 - x_i)}
    \\
    &\propto \theta_t^{\alpha_0 - 1 + \sum_{t} \zeta_i x_i} (1 - \theta_t)^{\beta_0 - 1 + \sum_t \zeta_i - x_i \zeta_i}.
\end{align}
So the MF-posterior is proportional to a beta distribution
\begin{equation}
\begin{aligned}
    \pi(\theta_t \given \mathbf{D}_{1:t}) &= \distBeta(\alpha_t, \beta_t),
    \\
    \alpha_t &\define \alpha_0 + \sum_{i=1}^t \zeta_i x_i,
    \\
    \beta_t &\define \beta_0 + \sum_{i=1}^t \zeta_i (1 - x_i).
\end{aligned}
\end{equation}
The MF-posterior predictive is:
\begin{align}
    &p(x_{t+1} \mid \zeta_{t+1}, \mathbf{D}_{1:t})
    \\
    &= \int_0^1 p_{\theta_t}(x_{t+1})^{\zeta_{t+1}} p(\theta_t \mid \mathbf{D}_{1:t}) \differential\theta_t
    \\
    &= \int_0^1 \left( \theta_t^{x_{t+1}} (1-\theta_t)^{1 - x_{t+1}} \right)^{\zeta_{t+1}} \left( \frac{1}{\text{B}(\alpha_t, \beta_t)} \theta_t^{\alpha_t - 1} (1 - \theta_t)^{\beta_t - 1} \right) \differential\theta_t
    \\
    &= \frac{1}{\text{B}(\alpha_t, \beta_t)} \int_0^1 \theta_t^{\zeta_{t+1} x_{t+1} + \alpha_t - 1} (1-\theta_t)^{\zeta_{t+1}(1 - x_{t+1}) + \beta_t - 1} \differential\theta_t
    \\
    &= \frac{\text{B}\!\left(
    \alpha_t + \zeta_{t+1} x_{t+1}, \beta_t + \zeta_{t+1} (1 - x_{t+1}) \right)}{\text{B}(\alpha_t, \beta_t)}.
\end{align}
The last step as, as in the general case, depends on knowing the normalizer of the beta distribution. Notice that the base measure $h_1(x_t)$ of the Bernoulli distribution is one, and therefore $[h_1(x_t)]^{\zeta_t} = 1$.

% ==============================================================================
\section{Alternative decision rule}
\label{app:alternative_decision_rule}
% ==============================================================================
%
\begin{figure*}[t]
\centering
\centerline{\includegraphics[width=0.5\textwidth]{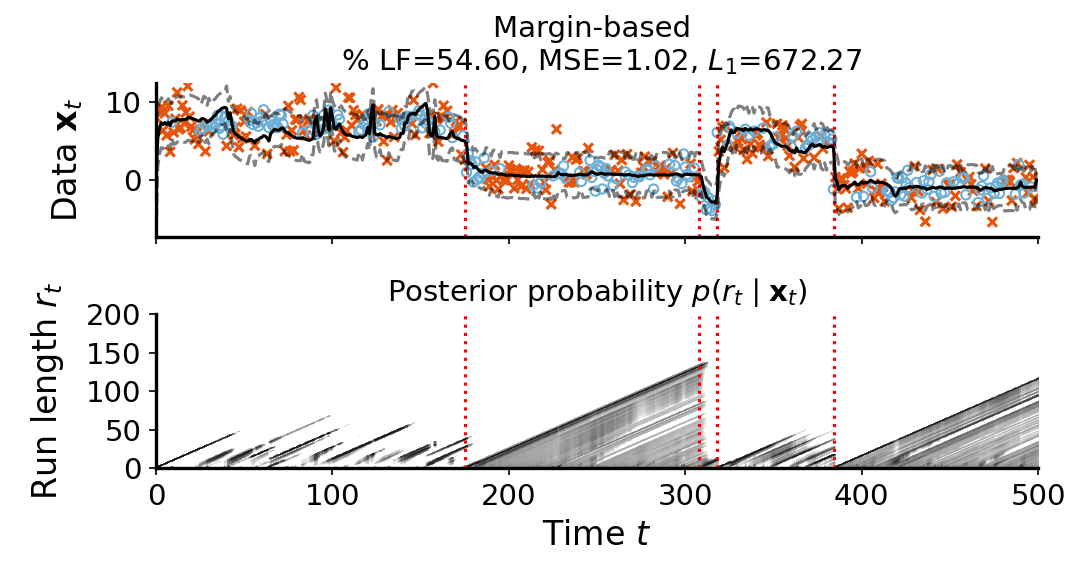}}
\caption{Orange x marks and blue circles denote low- and high-fidelity data respectively. A two-fidelity model that actively selects the lower fidelity if its information gain is close in value to the higher fidelity's information gain (\Cref{eq:alt_decision_rule}).}
\label{fig:margin_comparison}
\end{figure*}
Here, consider the scenario of just two fidelities, low fidelity $\zeta_{\textsf{low}}$ and high fidelity $\zeta_{\textsf{high}}$. An alternative decision rule to~\Cref{eq:decision_rule} would be to choose the lower fidelity when its utility or information gain is within some margin hyperparameter $\delta$ of the higher fidelity's utility:
\begin{equation}
    \zeta_t^{\star} = \begin{cases}
        \zeta_{\textsf{low}} & \text{if $| \mcU(\zeta_{\textsf{low}}) - \mcU(\zeta_{\textsf{high}})| < \delta$,}
        \\
        \zeta_{\textsf{high}} & \text{otherwise.}
    \end{cases}
    \label{eq:alt_decision_rule}
\end{equation}
However, we found that results on the Gaussian model in~\Cref{sec:numerical_experiments} were not promising (\Cref{fig:margin_comparison}). The model would frequently switch between fidelities because the utilties $\mcU(\zeta_{\textsf{low}})$ and $\mcU(\zeta_{\textsf{high}})$ were quite close in value. We found that information rate was more stable because it requires a more significant change in information gain to induce a switch.

% ==============================================================================
\clearpage
\section{MF-BOCD algorithm in didactic code}\label{app:didactic_code}
% ==============================================================================

This Python code is a didactic example of the MF-BOCD algorithm. At each time step, the algorithm (1) chooses a data fidelity using maximal information rate; (2) observes a datum of the chosen fidelity; (3-4) computes the posterior predictive and run-length posterior distributions; (5) updates the model parameters; and (6) makes a prediction. Please see the code repository\footnote{\url{https://github.com/princetonlips/mf-bocd}} for a complete example.

Note that in practice, each datum will be observed by evaluating an observation model in real-time. Here, for clarity, we simply index into a pre-initialized data array.

\begin{minted}[
    fontsize=\footnotesize,
    linenos]{python}
import numpy as np
from   scipy.special import logsumexp

def mf_bocd(data, model, hazard, costs):
    J, T        = data.shape
    log_message = np.array([1])
    log_R       = np.ones((T+1, T+1))
    log_R[0, 0] = 1
    pmean       = np.zeros(T)
    igs         = np.empty(J)
    choices     = np.empty(T)

    for t in range(1, T+1):
        
        # 1. Choose fidelity.
        rl_post = np.exp(log_R[t-1, :t])
        for j in range(J):
            igs[j] = compute_info_gain(t, model, rl_post, log_message, hazard, j)
        j_star = np.argmax(igs / costs)
        choices[t-1] = j_star

        # 2. Observe new datum.
        x = data[j_star, t-1]
        
        # 3. Compute predictive probabilities.
        log_pis = model.log_pred_prob(t, x, j_star)

        # 4. Estimate run length distribution.
        log_growth_probs = log_pis + log_message + np.log(1 - hazard)
        log_cp_prob      = logsumexp(log_pis + log_message + np.log(hazard))
        new_log_joint    = np.append(log_cp_prob, log_growth_probs)
        log_R[t, :t+1]   = new_log_joint
        log_R[t, :t+1]  -= logsumexp(new_log_joint)
    
        # 5. Update model parameters and message pass.
        model.update_params(t, x, j_star)
        log_message = new_log_joint
        
        # 6. Predict.
        pmean[t-1] = np.sum(model.mean_params[:t] * rl_post)

    return choices, np.exp(log_R), pmean
\end{minted}

% ==============================================================================
\pagebreak
\section{Ablation studies}\label{app:ablations}
% ==============================================================================

Here, we report the results of an ablation study for the multi-fidelity Gaussian and multi-fidelity Bernoulli models described in \Cref{sec:model_examples}. For varying costs, a multi-fidelity model using information gain-based switching was run on data generated from their respective data generating proceses. The percentage of low-fidelity observations was recorded; call this $P_{\textsf{low}}$. Then a randomized multi-fidelity model was run on the same dataset. At each time step, the randomized model chose low-fidelity data based on a Bernoulli random variable with bias $P_{\textsf{low}}$. The goal of this experiment is to demonstrate that when the model switches to high-fidelity data is important to model performance, not just the fact that some percentage of high-fidelity data are used. We found that for both Gaussian (\Cref{tab:ablation_gaussian}) and Bernoulli data (\Cref{tab:ablation_bernoulli}), choosing when to switch fidelities was often useful.

\begin{table}[h]
\caption{Ablation study for multi-fidelity Gaussian models. ``LF only'' is BOCD using only low-fidelity data. Mean and two standard errors, representing 95\% confidence intervals, are reported over 200 trials. Bold numbers indicate statistically significant using 95\% confidence intervals.}
\label{tab:ablation_gaussian}
\centering
% \resizebox{\textwidth}{!} {%
\begin{tabular}{c|ccc|ccc}
& \multicolumn{3}{c}{MSE} & \multicolumn{3}{c}{$L_1$}
\\
\toprule
LF (\%) & LF only & Random & Info-based & LF only & Random & Info-based \\
\midrule
1 & \multirow{8}{*}{0.879 (0.034)} & 0.046 (0.056) & 0.003 (0.001) & \multirow{8}{*}{270.87 (8.35)} & \textbf{5.98} (3.06) & 73.92 (9.61) \\
2 & & 0.125 (0.073) & 0.111 (0.046) & & \textbf{18.18} (7.37) & 77.68 (9.79) \\
38 & & 0.680 (0.118) & \textbf{0.494} (0.059) & & 162.31 (12.97) & 161.05 (11.08) \\
53 & & 0.702 (0.091) & \textbf{0.483} (0.066) & & 183.40 (11.36) & 173.01 (10.46) \\
60 & & 0.752 (0.140) & \textbf{0.452} (0.037) & & 186.11 (10.89) & 174.95 (10.13) \\
67 & & 0.665 (0.075) & \textbf{0.466} (0.036) & & 187.72 (10.01) & 173.41 (9.91) \\
74 & & 0.643 (0.064) & \textbf{0.480} (0.043) & & 182.18 (9.48) & 175.88 (9.36) \\
80 & & 0.656 (0.087) & \textbf{0.492} (0.044) & & 184.66 (9.20) & 175.70 (9.20) \\
97 & & 0.547 (0.028) & 0.537 (0.028) & & 176.76 (9.40) & 175.34 (9.33) \\
\bottomrule
\end{tabular}
% }
\end{table}

\begin{table}[h]
\caption{Ablation study for multi-fidelity Bernoulli models. ``LF only'' is BOCD using only low-fidelity data. Mean and two standard errors, representing 95\% confidence intervals, are reported over 200 trials. Bold numbers indicate statistically significant using 95\% confidence intervals.}
\label{tab:ablation_bernoulli}
\centering
% \resizebox{\textwidth}{!} {%
\begin{tabular}{c|ccc|ccc}
& \multicolumn{3}{c}{MSE} & \multicolumn{3}{c}{$L_1$}
\\
\toprule
LF (\%) & LF only & Random & Info-based & LF only & Random & Info-based \\
\midrule
9 & \multirow{8}{*}{0.123 (0.009)} & 0.003 (0.001) & \textbf{0.002} (0.000) & \multirow{8}{*}{186.27 (7.02)} & 45.55 (6.04) & 40.31 (5.43) \\
21 & & 0.008 (0.001) & 0.009 (0.002) & & 76.42 (7.68) & 71.88 (7.54) \\
25 & & 0.011 (0.002) & 0.011 (0.002) & & 84.47 (8.11) & 80.61 (7.89) \\
46 & & 0.025 (0.003) & 0.021 (0.003) & & 124.80 (7.07) & 117.34 (7.31) \\
61 & & 0.040 (0.004) & 0.034 (0.005) & & 143.87 (6.34) & 139.88 (7.23) \\
68 & & 0.050 (0.005) & \textbf{0.040} (0.005) & & 158.48 (6.34) & 149.79 (7.02) \\
73 & & 0.057 (0.005) & 0.048 (0.006) & & 163.68 (6.39) & 158.08 (6.66) \\
83 & & 0.077 (0.006) & \textbf{0.064} (0.007) & & 174.01 (6.21) & 170.26 (6.49) \\
90 & & 0.098 (0.007) & \textbf{0.082} (0.007) & & 184.46 (6.11) & 178.86 (6.28) \\
\bottomrule
\end{tabular}
% }
\end{table}

% ==============================================================================
\clearpage
\section{Experimental details}\label{app:experimental_details}
% ==============================================================================

% ==============================================================================
\subsection{CamVid experiments}
% ==============================================================================

The pretrained MobileNets were downloaded from the Fastseg Python library.\footnote{\url{https://github.com/ekzhang/fastseg}}

We can estimate the computational cost of MF-BOCD ($\lambda_{\textsf{MF}}$) relative to BOCD using only high- ($\lambda_{\textsf{HF}}$) and low- ($\lambda_{\textsf{LF}}$) fidelity data. We used 85 low- and 86 high- fidelity observations. The low- (high-) fidelity observation model required 19.48 (36.89) billion flops (\Cref{tab:mobile_micro_nets}). Computing the information gain required 465,291 flops. The total cost of our algorithm in billions of flops is
\begin{align*}
    \lambda_{\textsf{LF}} &= 171 \!\times\! 19.5 \approx 3333,
    \\
    \lambda_{\textsf{HF}} &= 171 \!\times\!36.9 \approx 6303,
    \\
    \lambda_{\textsf{MF}} &= 0.00046 \!+\! (85 \!\times\! 19.5) + (86 \!\times\! 36.7) \approx 4827.
\end{align*}
As we can see, decision-making has a marginal cost.

\begin{table}[h]
\caption{Observation model details for CamVid and MIMII experiments. (CamVid) The high-fidelity model has roughly twice times the number of flops and higher accuracy as measured by intersection-over-union (IoU) on the Cityscapes dataset. (MIMII) The high-fidelity model requires roughly 250 times as many floating point operations (ops). ``FC'', ``M'', and ``B'' mean fully-connected, millions, and billions respectively.}
\label{tab:mobile_micro_nets}
\centering
\resizebox{0.6\columnwidth}{!}{%
\begin{tabular}{l|cccc}
\toprule
\midrule
& Fidelity & Model & Ops & Accuracy
\\
\midrule
\multirow{2}{*}{CamVid} & HF & V3-large & 36.86 B & 72.3 (IoU\%)
\\
& LF & V3-small & 19.48 B & 67.4 (IoU\%)
\\
\midrule
\multirow{2}{*}{MIMII} & HF & MicroNet-AD(M) & 124.7 M & 96.15 (AUC\%)
\\
& LF & Two-layer FC & 0.5 M & 86.7 (AUC\%)
\\
\bottomrule
\end{tabular}
}
\end{table}

\begin{figure*}[tbh]
\centering
\centerline{\includegraphics[width=0.6\textwidth]{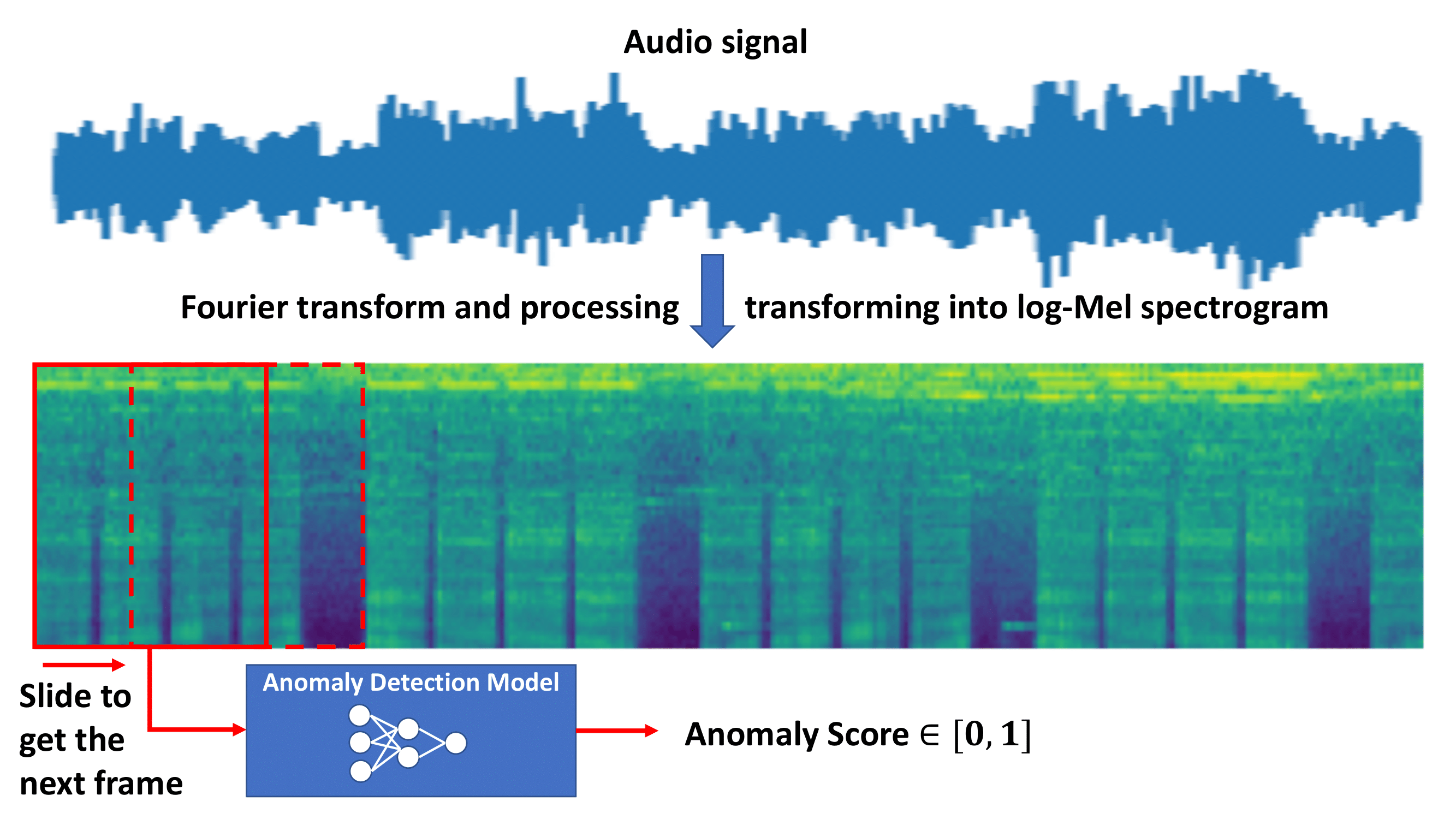}}
\caption{Illustration of pipeline to generate anomaly scores from log-Mel spectrograms using deep neural networks.}
\label{fig:anomaly_detection_illustration}
\end{figure*}

% ==============================================================================
\subsection{MIMII experiment}
% ==============================================================================

\begin{figure*}[t]
\vskip 0.2in
\begin{center}
\centerline{\includegraphics[width=0.8\textwidth]{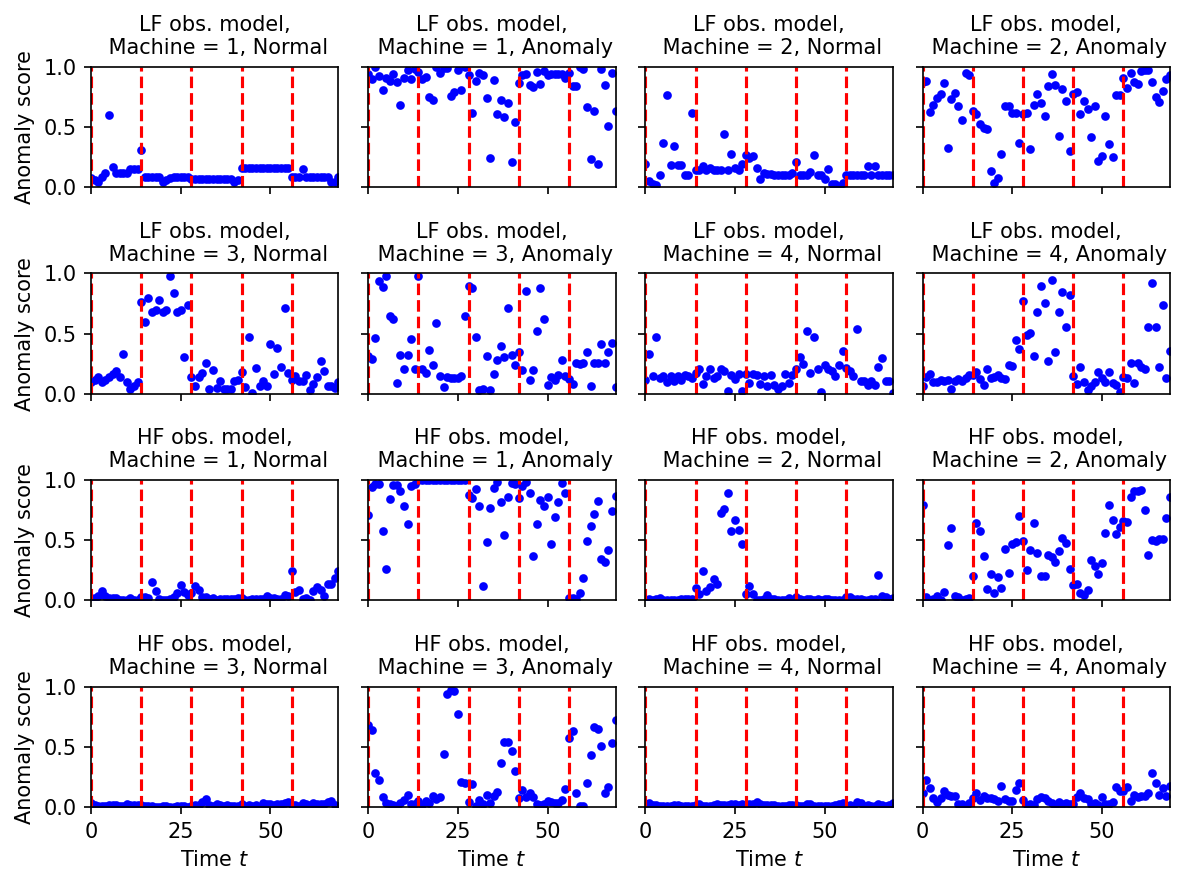}}
\caption{Examples of MIMII anomaly scores, five audio clips for each machine. Dashed red lines separate audio clips.}
\label{fig:mimii_data_examples}
\end{center}
\vskip -0.2in
\end{figure*}

\begin{figure*}[t]
\vskip 0.2in
\begin{center}
\centerline{\includegraphics[width=\textwidth]{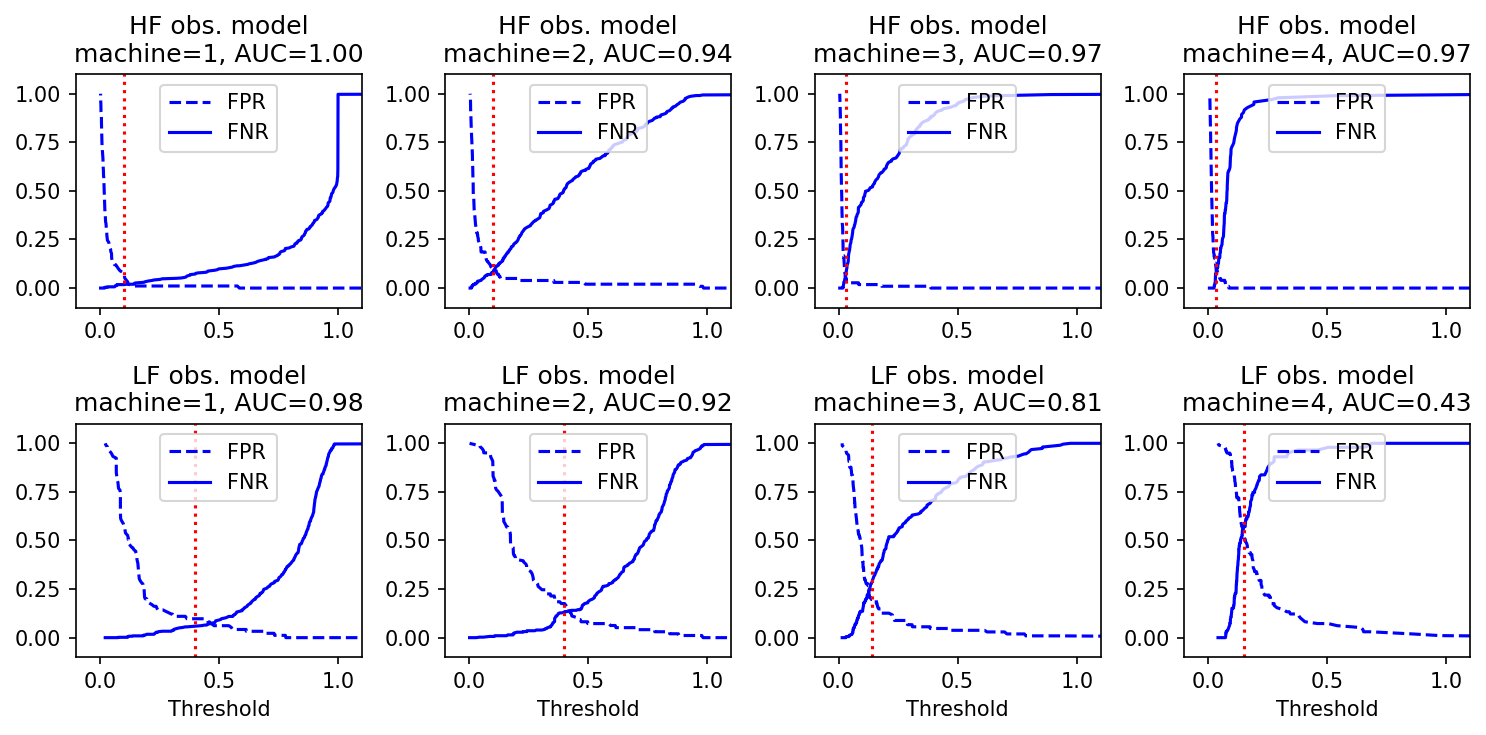}}
\caption{False positive (FPR) and false negative rates (FNR) for high- (HF) and low- (LF) fidelity observation models on MIMII cross-validation data. Vertical dashed lines indicated the chosen threshold}
\label{fig:mimii_threshold_estimation}
\end{center}
\vskip -0.2in
\end{figure*}

In the MIMII experiment, the output of the observation models (\Cref{tab:mobile_micro_nets}) is a scalar anomaly score in the range $[0, 1]$, where 0 indicates normal machine operation. An illustration of how these scores are obtained for an audio clip is shown in \Cref{fig:anomaly_detection_illustration}. To convert these anomaly scores to binary numbers for a Bernoulli multi-fidelity posterior predictive model, we thresholded the scores to integers in $\{0, 1\}$. The quality of the observation models depends on the choice of threshold. For examples of these data, see \Cref{fig:mimii_data_examples}. To select the appropriate threshold, we used the intersection of the false negative and false positive rate curves, which corresponds to the top-left corner of the receiver operating characteristic (ROC) curves for each machine and each observation model (\Cref{fig:mimii_threshold_estimation}).

\begin{figure}[t]
\centering
\centerline{\includegraphics[width=\columnwidth]{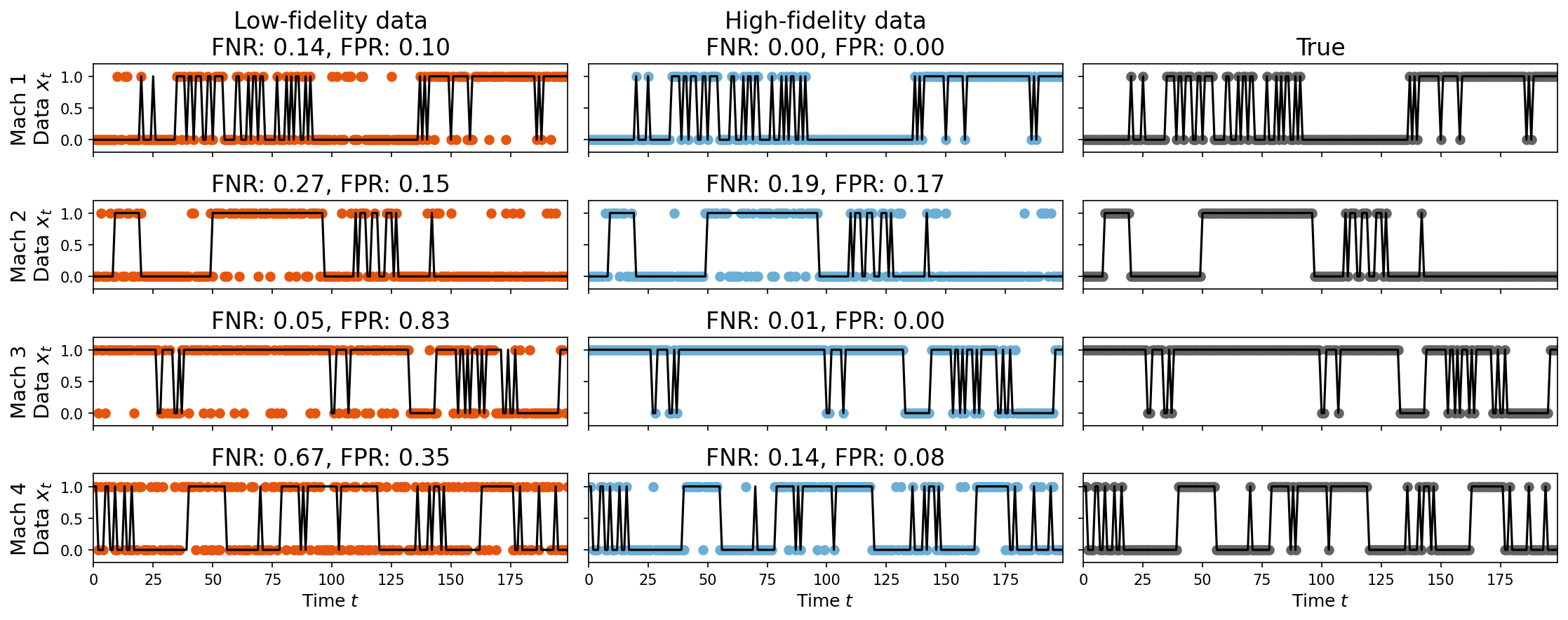}}
\caption{Illustration of MIMII data after converting log-Mel spectrograms to binary numbers with machine- and observation model-specific thresholds. The true binary value is denoted with a black line.}
\label{fig:machine_comparison}
\end{figure}

\end{appendices}

\end{document}